\pdfoutput=1

\documentclass[11pt]{article}

\usepackage{acl}

\usepackage{times}
\usepackage{latexsym}

\usepackage[T1]{fontenc}

\usepackage[utf8]{inputenc}

\usepackage{microtype}

\usepackage{graphicx} 
\usepackage{float} 
\usepackage{subfigure}
\usepackage{tabularx}
\usepackage{array}
\usepackage{booktabs}
\usepackage{amsmath}
\usepackage{amsfonts}
\usepackage{arydshln}
\usepackage[ruled,linesnumbered]{algorithm2e}
\title{Unsupervised Question Answering via Answer Diversifying}

\author{Yuxiang Nie,~~Heyan Huang\thanks{~~Corresponding author},~~Zewen Chi,~~Xian-Ling Mao \\
        Beijing Institute of Technology \\ \texttt{\{nieyx,hhy63,czw,maoxl\}@bit.edu.cn}}

\begin{document}
\maketitle
\begin{abstract}
Unsupervised question answering is an attractive task due to its independence on labeled data. Previous works usually make use of heuristic rules as well as pre-trained models to construct data and train QA models. However, most of these works regard named entity (NE) as the only answer type, which ignores the high diversity of answers in the real world. To tackle this problem, we propose a novel unsupervised method by diversifying answers, named \textbf{DiverseQA}. Specifically, the proposed method is composed of three modules: data construction, data augmentation and denoising filter. Firstly, the data construction module extends the extracted named entity into a longer sentence constituent as the new answer span to construct a QA dataset with diverse answers. Secondly, the data augmentation module adopts an answer-type dependent data augmentation process via adversarial training in the embedding level. Thirdly, the denoising filter module is designed to alleviate the noise in the constructed data. Extensive experiments show that the proposed method outperforms previous unsupervised models on five benchmark datasets, including SQuADv1.1, NewsQA, TriviaQA, BioASQ, and DuoRC. Besides, 
the proposed method shows strong performance in the few-shot learning setting.\footnote{We have released our codes and data in \url{https://github.com/JerrryNie/DiverseQA}.}
\end{abstract}

\section{Introduction}
\begin{figure}[t]
    \begin{minipage}{\linewidth}
    \centering
        \includegraphics[width=.9\linewidth]{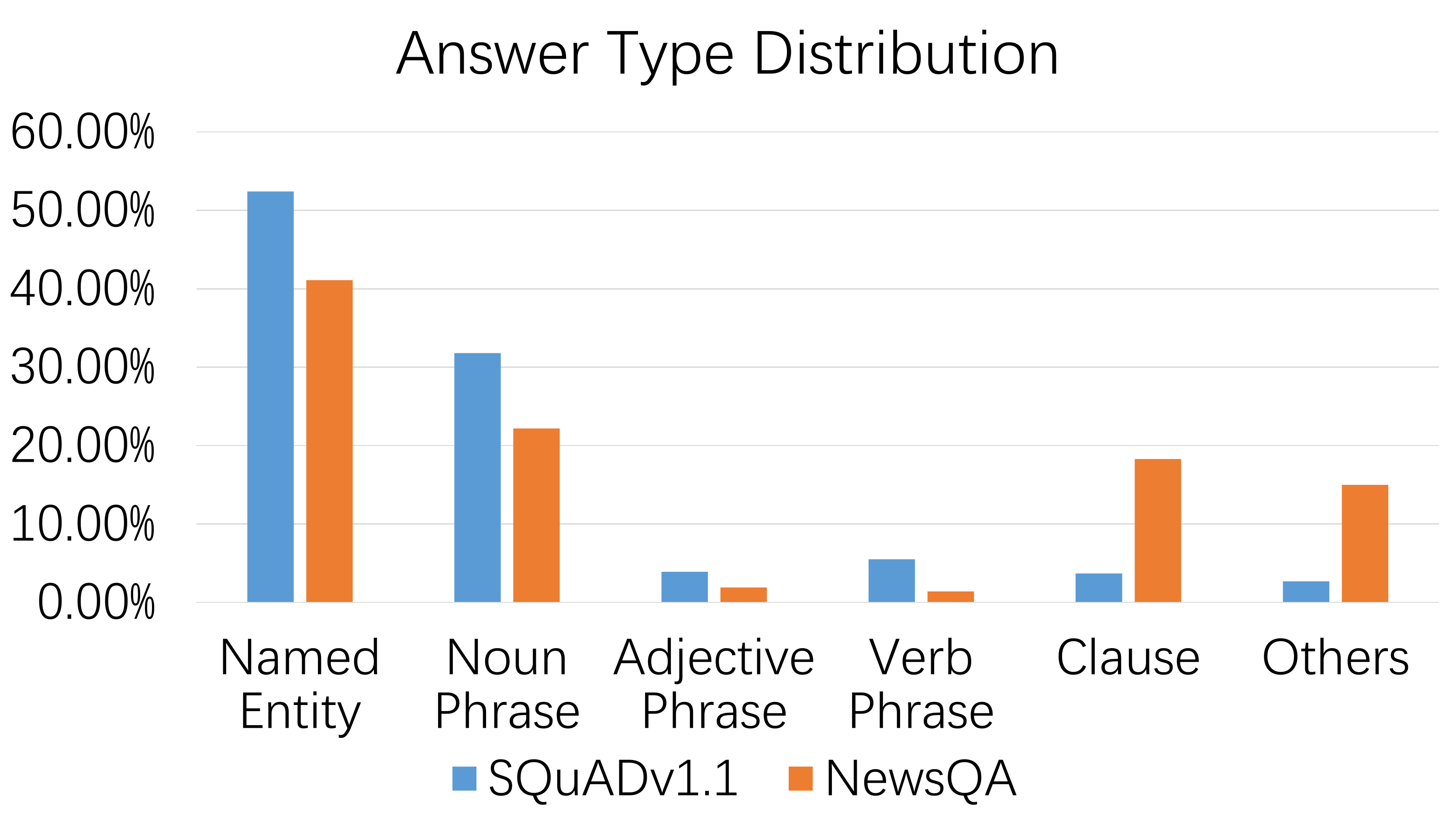}\\
        \vspace{0.02cm}
    \end{minipage}
    \centering
    \vspace{-0.2cm}
    \caption{The answer type distributions of SQuADv1.1 and NewsQA, where we notice that named entities are just a fraction of each dataset.}
    \label{fig:answer-type-distribution-graph}
\end{figure}
Extractive question answering (extractive QA) aims to provide answer spans extracted from the context to answer questions. It can improve the interaction between humans and machines in applications such as search engines and dialog systems.

Existing extractive QA methods can be divided into two categories: supervised QA and unsupervised QA. Traditionally, for supervised QA \cite{seo2016bidirectional}, human-labeled context, questions and answers are given to train a QA model. Since the construction cost of labeled data is too high for supervised QA, recently, researchers pay more attention to the unsupervised setting, where QA pairs are unavailable. For example, \citet{lewis2019unsupervised} propose an unsupervised machine translation model to generate QA pairs.
\citet{li2020harvesting}, \citet{hong2020handling} and \citet{lyu2021improving} try to alleviate the overlap between the context and the generated question.
However, most existing works regard named entities as the only answer type, which ignores the high diversity of answers in the real world. For instance, as shown in Figure \ref{fig:answer-type-distribution-graph}, in SQuADv1.1 \cite{rajpurkar2016squad} and NewsQA \cite{trischler2017newsqa}, the answer type of named entities only account for 52.4\% and 41.1\% respectively.

\begin{figure}[t]
\setlength{\belowcaptionskip}{-0.5cm}
  \centering
  \includegraphics[width=.9\linewidth]{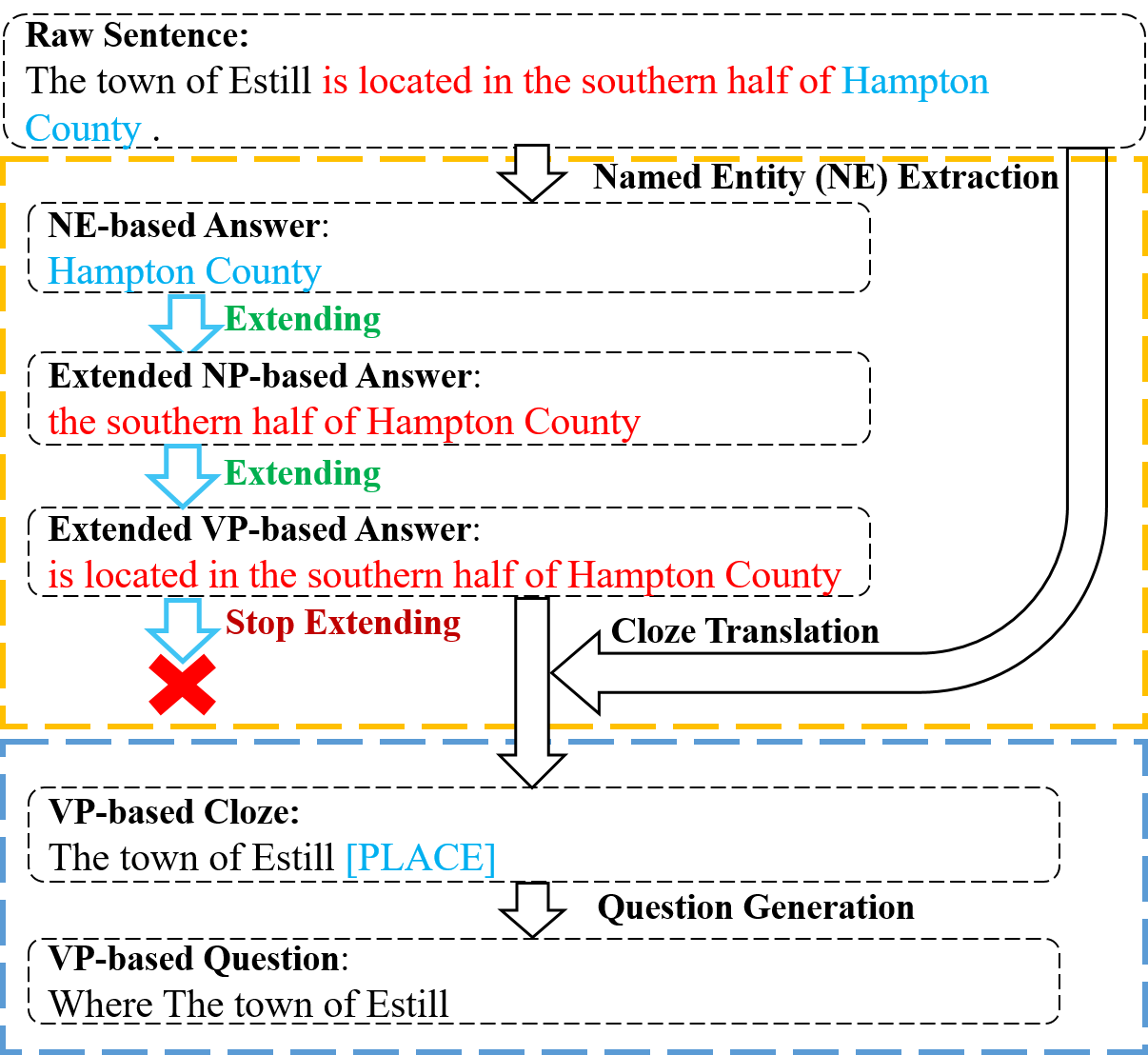}
  \caption{An example of our data constructing module. We extract the named entity (NE) in the \textit{Raw Sentence} and extend it into a longer sentence constituent until meeting the stop extending condition. The final extended answer span is a VP in this example, which is used in cloze translation for question generation.}
  \label{fig:answer-extension}
\end{figure}

To solve the problem, an intuitive way is to extract each type of answer spans independently in the raw text. Yet, it leads to two critical problems. Firstly, it's hard to determine the proportion of each answer type in the synthetic dataset. Secondly, extracting answers without any guidance could probably generate trivial and noisy question-answer pairs. To tackle the problem above, we propose \textbf{DiverseQA}, a novel unsupervised QA method. Specifically, the proposed method consists of three modules: data construction, data augmentation and denoising filter. Firstly, the data construction module employs a simple answer-extending rule to construct a dataset with diverse answers. As shown in Figure \ref{fig:answer-extension}, a named entity (NE) is extracted and extended into a longer answer span, which should be a constituent of the sentence (e.g. noun phrase, verb phrase). In this way, the proportion of each type of answers can be obtained entirely from the original text. Besides, NE-based extension largely guarantees 
that extracted answers are meaningful.
Secondly, an answer-type dependent data augmentation module is proposed. Concretely, an adjusting vector generator is designed to produce answer-type enhanced QA pairs in the embedding level. Then, a discriminator classifies the embeddings into the corresponding answer type while minimizing the KL divergence between the distribution and its prior to fool the discriminator in an adversarial way. Thirdly, the denoising filter is applied to alleviate the noise in the synthetic QA pairs.

Extensive experiments on six benchmarks, including SQuADv1.1 \cite{rajpurkar2016squad}, TriviaQA \cite{joshi2017triviaqa}, NaturalQuestions \cite{kwiatkowski2019natural}, NewsQA \cite{trischler2017newsqa}, BioASQ \cite{tsatsaronis2015overview} and DuoRC \cite{saha2018duorc} show that DiverseQA outperforms previous unsupervised methods on five datasets and obtains comparable results on one dataset among unsupervised QA models. Further analysis shows that DiverseQA can largely improve the question-answering ability of the model on diverse answer types. In addition, our method shows strong performance in the few-shot learning setting.

The contributions of our method are as follows:
\begin{itemize}
\setlength{\itemsep}{0pt}
\setlength{\parsep}{0pt}
\setlength{\parskip}{0pt}
    \item We propose DiverseQA, a novel method to improve an unsupervised QA model to handle answers beyond named entities.
    \item Our method outperforms previous unsupervised works on five benchmarks and reaches the comparable result on a benchmark.
    \item Further analysis shows that the drift of answer length distribution and the quality of extracted answers are important to the performance of the model.
\end{itemize}

\section{Related Work}
\paragraph{Data Augmentation.} Data augmentation methods can be regarded as regularizers to make the model robust and reduce dependence on the training data. In computer vision domains \cite{krizhevsky2012imagenet}, geometric transformation and color space transformation are effective. In natural language processing, word removing, synonym replacing and back-translation can enlarge the diversity of examples \cite{xie2020unsupervised}. \citet{lee2021learning} proposes an embedding-level data augmentation method to improve the performance of QA on out-of-distribution data. This work relies on supervised (low noise) training instances, while in the unsupervised (high noise) scenario, embedding-level data augmentation methods have not been explored yet.

\paragraph{Extractive Question Answering.} Extractive question answering (extractive QA) aims to output an answer span given the context (containing the answer) and the question. It has gained much progress with the help of large labeled datasets such as SQuAD, NewsQA, and TriviaQA. To better handle these datasets, strong extractive QA models are proposed, including BiDAF \cite{seo2016bidirectional}, BERT \cite{devlin2018bert} and RoBERTa \cite{liu2019roberta}. However, because they largely depend on human-annotated data, the lack of labeled data for some specific domains constrains the capacity of the supervised extractive QA model.

\paragraph{Unsupervised Question Answering.} Unsupervised question answering (unsupervised QA) becomes attractive among researchers recently \cite{lewis2019unsupervised}. Like supervised extractive QA, given context and question, the model needs to extract a text span from the context to answer the question. The difference is that in the unsupervised setting, models need to learn to answer the question without any human-labeled $\langle\text{context},\ \text{question},\ \text{answer} \rangle $ triples, which is a more challenging task than the supervised counterpart. \citet{lewis2019unsupervised} extracts named entity from the context, and then trains an unsupervised neural machine translation model to convert the cloze question into a natural question. \citet{li2020harvesting} makes use of cited documents and a refine phase to alleviate serious question-context overlapping and improve answer diversity. \citet{hong2020handling} uses paraphrasing and trimming to remove unanswerable questions as well as alleviate the context-question similarity problem. \citet{lyu2021improving} takes advantage of a supervised summarization dataset to tackle the context-question overlapping problem. However, most of the existing works regard the named entity (NE) as the only answer type, while other types (e.g. noun phrases, adjective phrases, verb phrases, sub-clauses) are nearly ignored. Although in \citet{lewis2019unsupervised}, the inclusion of noun phrases (NPs) was discussed, the reported poor performance eventually led the author to use the NE-based synthetic QA pairs. In \citet{li2020harvesting}, even though generating more answers is considered, most of the model-generated answers have strong relationships with the named entity and we will discuss it later in Section \ref{sec:effects_of_diverse_answer_types}.

\section{Method}

To explore answer types beyond named entities,
we propose an unsupervised method \textbf{DiverseQA}, which can be divided into three modules, data augmentation, data construction and denoising filter. Firstly, a simple but effective span extending rule is applied in the data construction module to do answer extraction and natural question generation to construct a synthetic QA dataset with diverse answers. Secondly, an answer-type dependent data augmentation module via adversarial training is designed to produce high-quality augmented QA pairs. Thirdly, a denoising filter is proposed to alleviate high noise in the dataset. In the 
next few subsections, we will introduce the three modules in detail.


\subsection{Synthetic QA Data Construction via Span Extension}

In this section, we illustrate our QA data constructing module, which can be divided into two steps. Firstly, we extract answers in the text through span extension. Secondly, we take advantage of pseudo-NER label to generate questions.

\subsubsection{Answer Generation via Span Extension}
As shown in Figure \ref{fig:answer-extension}, given a \textit{Raw Sentence}, first of all, we extract the named entity span \textit{Hampton County}. Then, we extract all the constituents containing \textit{Hampton County}. We extend the span into a longer constituent iteratively. When the next extending span makes up more than $\omega$\% of the \textit{Raw sentence}, we stop extending and regard the extended span as the final answer span. In this example, the NE is firstly extended into a longer NP. Again, the NP is extended into the VP. As the VP `\textit{is located in the southern half of Hampton County}' is the longest constituent while making up less than $\omega$\% (called \textit{Span Extending Threshold}) of the \textit{Raw Sentence}, it becomes the answer span of the QA pair. The intuition is that even if many constituents are trivial, the constituent containing the named entity and making up a proper portion of the \textit{Raw Sentence} might be meaningful to create a high-quality QA pair.

\subsubsection{Question Generation with Pseudo-NER Label}

To construct a NE-based QA pair, the NER label of the answer span is mapped to a question word \cite{lewis2019unsupervised} for generating the natural question. However, it cannot be directly applied to other answer types. To tackle the problem, we regard the original NER label of the named entity as the \textit{pseudo-NER label} of the extended answer span. After that, we replace the answer span with the high-level NER mask token \cite{lewis2019unsupervised}. The intuition is that the semantic information of the extracted named entity can be probably consistent with the extended constituent.

Followed \citet{lewis2019unsupervised} and \citet{li2020harvesting}, we do the mask token mapping and cloze to natural question conversion.
\vspace{-0.1cm}
\subsection{Answer-type Dependent Data Augmentation}
We firstly introduce the background of extractive QA, and then illustrate how we design our answer-type dependent data augmentation modules.
\vspace{-0.1cm}
\subsubsection{Backgrounds of Extractive Question Answering}
Given question $\mathbf{q}=(q_1,q_2,...,q_m)$ and context $\mathbf{c}=(c_1,c_2,...,c_n)$, extractive QA models aim to predict the start and end token of the answer span $\mathbf{a}=(a_1,a_2)$ from the context. Assuming that there are $N$ observations: $\{\mathbf{c}^{(i)},\mathbf{q}^{(i)},\mathbf{a}^{(i)}\}_{i=1}^N$, we estimate the model parameter $\theta$ by maximizing the following function:
\vspace{-0.4cm}
\begin{equation}
    \mathcal{L}_{\text{MLE}}(\theta)=\sum_{i=1}^Np_{\theta}(\mathbf{a}^{(i)}|\mathbf{c}^{(i)},\mathbf{q}^{(i)})
\end{equation}

\subsubsection{Answer-type Dependent Embedding Adjustment}
\label{sec:embedding-adjusting}
As QA instances might have specific feature related to answer types, motivated by \citet{lee2021learning}, we design an answer-type dependent embedding-level \textit{Adjusting Vector} to create high quality instances for model training.

 The proposed adjusting vector $\mathbf{z}\in\mathbb{R}^{d\times (m+n+r)}$ is sampled from the distributions $q_{\phi}(\mathbf{z}|\mathbf{x},l)$ to augment the input sequence $\mathbf{x}\in\mathbb{R}^{d\times (m+n+r)}$, where $r$ is the special token length, $d$ is the size of a word embedding vector, $\mathbf{x}$ is the embeddings, $l$ is an answer type. We use the element-wise production between the embedding and the Adjusting Vector as extra data to train a QA model by maximizing the log-likelihood function of $p_{\theta}(\mathbf{a}|\mathbf{x,z})$:
\begin{equation}
\label{eqn:swep}
\begin{aligned}
\mathcal{L}_{\text{Adjust}}(\mathbf{\theta,\phi})=&\sum_{i=1}^N\mathbb{E}_{q_\phi(\mathbf{z}|\mathbf{x},l)}\left[\text{log}\ p_{\theta}(\mathbf{a}^{(i)}|\mathbf{x}^{(i)}, \mathbf{z})\right]\\
+&\beta \text{KL}\left(q_{\phi}(\mathbf{z}|\mathbf{x}^{(i)},l)||p_{\psi}(\mathbf{z})\right)
\end{aligned}
\end{equation}
where $p_{\psi}(\mathbf{z})$ is the prior of the distribution $q_{\phi}(\mathbf{z}|\mathbf{x},l)$, $\psi$ is the predetermined parameter of the distribution. We assume it obeys multivariate Gaussian distributions $\mathcal{N}(\mathbf{1}, \gamma\mathbf{I}_{\text{d}})$. 

\begin{figure}[t]
\setlength{\belowcaptionskip}{-0.5cm}
  \centering
  \includegraphics[width=.9\linewidth]{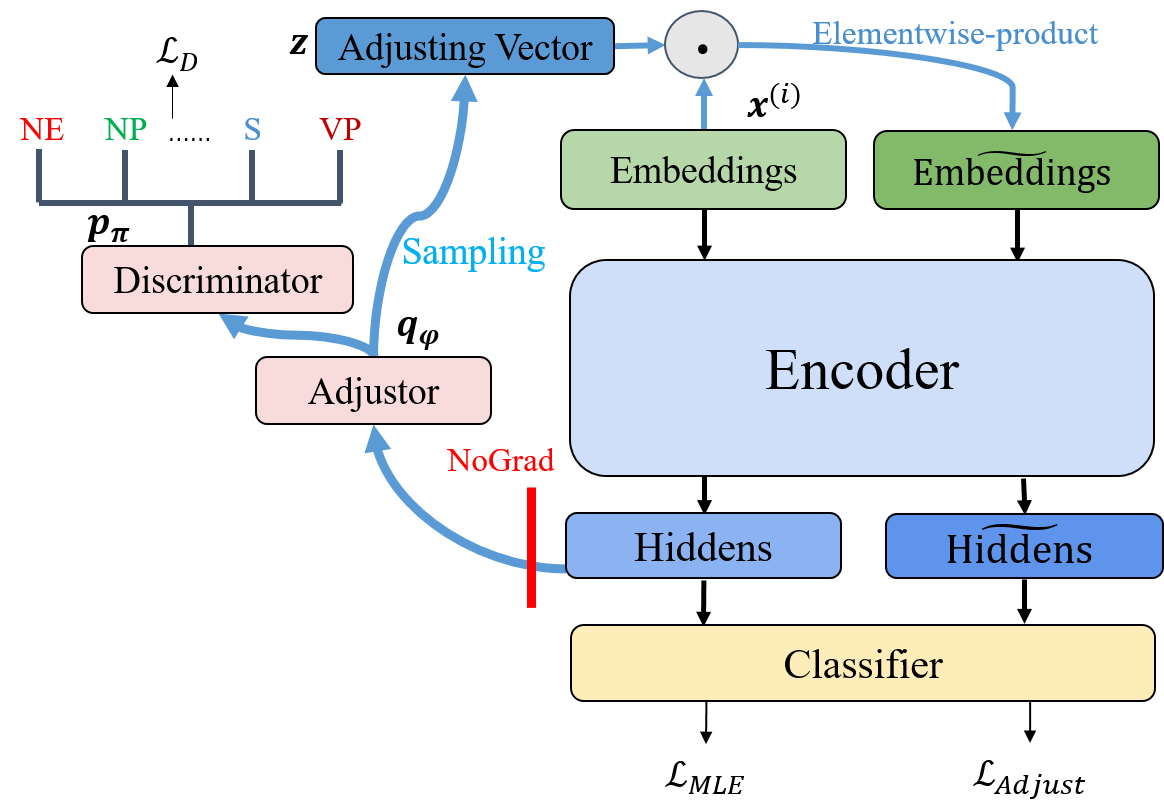}
  \caption{Our data augmentation module. The Embeddings of the input sequence are firstly encoded into the Hiddens. Then, the Adjustor module receives the Hiddens and produces an Adjusting Vector to adjust the Embeddings by the element-wise product. Meanwhile, the Discriminator classifies the produced vector into an answer type to make the vector answer-type dependent.}
  \label{fig:model}
\end{figure}


To make the embeddings answer-type dependent, we train a neural network as the \textit{discriminator} to classify the adjusting vector into the corresponding answer type. The discriminator receives the adjusting vector $\mathbf{z}_j\in\mathbb{R}^d$ and classifies the vector into a corresponding answer type by using the predicting distribution $p_{\pi}(l|\mathbf{z}_j)=\frac{e^{f_l(\mathbf{z}_j)}}{\sum_{i=1}^Le^{f_i(\mathbf{z}_j)}}$, where $f_i(\mathbf{z}_j)$ denotes the logit of answer type label $i$ given $\mathbf{z}_j$. Due to imbalanced label distribution among different answer types, followed \citet{menon2020long}, we modify the distribution as: $p_{\pi}'(l|\mathbf{z}_j)=\frac{e^{f_l(\mathbf{z}_j)+\text{log}\ p_l}}{\sum_{i=1}^Le^{f_i(\mathbf{z}_j)+\text{log}\ p_i}}$, where $p_i$ is the frequency of the answer type $i$ and $\pi$ is the parameters of the discriminator. The discriminator is trained via optimizing the following loss:
\begin{equation}
    \mathcal{L}_{D}(\pi|\mathbf{z})=\sum_{i=1}^N\sum_{j=1}^{P}\sum_{l=1}^Ly_{il}p_{\pi}'(l|z_j^{(i)})
\end{equation}
where $P$ is the length of the input sequence, $L$ is the number of answer types. $y_{il}=1$ when the $i$-th observation is related to the answer type $l$, otherwise $y_{il}=0$.
To fool the discriminator, the Adjustor needs to minimize the KL divergence between its distribution $q_{\phi}$ and the prior $p_{\psi}$ in Eqn. \ref{eqn:swep}.
\paragraph{The Final Objective Function.} The final objective function is as follows:
\begin{equation}
\begin{aligned}
\mathcal{L}(\theta, \phi, \pi)&=\mathcal{L}_{\text{MLE}}(\theta) + \mathcal{L}_{\text{Adjust}}(\theta, \phi)+ \alpha\mathcal{L}_{D}(\pi)
\end{aligned}
\end{equation}
where $\alpha$ is the hyperparameter weighting between the question-answering predicting loss and the answer-type discriminating loss.


\begin{table*}[t]   
\fontsize{10.7}{11}\selectfont
    \centering
    \begin{tabular}{lcccccc}    
        \toprule    
        & {SQuADv1.1}& {TriviaQA} & {NQ} & {NewsQA} & {BioASQ} & {DuoRC}\\
        Models & EM/F1& EM/F1  & EM/F1 & EM/F1& EM/F1& EM/F1\\
        \midrule
        \multicolumn{7}{l}{\textit{Supervised Models}}\\
        \ \ \ Match-LSTM &  64.1/73.9& -/-& -/-&  -/-& -/-& -/-\\
        \ \ \ BiDAF &  66.7/77.3& -/-& -/-&  -/-& -/-& -/-\\
        \ \ \ BERT-base &  81.2/88.5& 69.4/74.3${}^\triangleleft$& 66.1/77.9${}^\triangleleft$&  49.4/64.4${}^\triangleleft$& -/-& -/-\\
        \ \ \ BERT-large &  84.2/91.1& 75.7/80.2${}^\triangleleft$& 69.0/80.8${}^\triangleleft$&  56.0/71.0${}^\triangleleft$& -/-& -/-\\
        \multicolumn{7}{l}{\textit{Trained with Supervised Summarization Dataset}}\\
        \ \ \ \citet{lyu2021improving} &  65.6/74.5&  36.7/43.0& 46.0/53.5&37.5/50.1& 32.0/43.2& 38.8/46.5\\
        \multicolumn{7}{l}{\textit{Unsupervised Models}}\\
        \ \ \ \citet{lewis2019unsupervised} &  44.2/54.7& 19.1/23.8$\dagger$& 27.5/35.1$\dagger$&  19.6/28.5$\ddagger$& 18.9/27.0$\dagger$& 26.0/32.6$\dagger$\\
        \ \ \ RefQA  &  62.5/72.6&  48.6/58.2$\ddagger$& \textbf{43.4}/55.7$\ddagger$&33.6/46.3& 42.5/58.9$\ddagger$& 38.0/49.4$\ddagger$\\
        \ \ \ DiverseQA &  \textbf{67.6/76.9}&  \textbf{52.5/60.8}& 41.3/\textbf{56.3}&\textbf{37.5/51.3}& \textbf{47.2/61.4}& \textbf{46.9/56.3}\\
        \bottomrule   
    \end{tabular}  
    \caption{\label{tab:compare_baselines} Results (EM/F1) of our method and various baselines on six different datasets. `$\dagger$' denotes results taken from \protect\citet{lyu2021improving}. `$\ddagger$' denotes results from our reimplementation of RefQA\protect\cite{li2020harvesting}. `$\triangleleft$' denotes our fine-tuned BERT on supervised data. Because the pre-processed training sets of BioASQ and DuoRC in MRQA are not released, we don't fine-tune BERT on them.}
\end{table*}

\subsection{Denoising Filter}\label{sec:denoising-filter}
We design a denoising filter to further alleviate the negative effect of the noise in the data. The filter is composed of the \textit{Top-K Filter} and the \textit{Substring Filter}. To apply them, we firstly use the QA model to do inference on the unseen synthetic data. Then, if the synthetic answer falls in the $K$ answers with the highest probability the model predicts (Top-K Filter) or the predicted instance is a substring of a NE-based answer span with predicting probability higher than $\gamma$ (Substring Filter), we keep it. Otherwise, we remove the instance. Then, we use the filtered data to train our fine-tuned model. In this process, Top-K Filter aims to choose QA pairs with high confidence as low noise instances. Substring Filter keeps extra NE-based QA pairs with high predicting probabilities for training. The idea comes from an observation that a substring of a NE can
probably represent the NE. For example, in the sentence \textit{Apple CEO Tim Cook introduces two new products}, “Tim Cook” is a NE referred to a specific person. Besides, “Tim” or “Cook” can also refer to him. Therefore, when the model predicts a substring of a NE with a high probability, it might refer to the original NE as the answer.

\section{Experiments}
\subsection{Experiment Setup}
\paragraph{Unsupervised QA Dataset Construction.} We use Wikiref \cite{li2020harvesting} as the original text to construct QA pairs.

To extract answer spans, firstly, we use Spacy\footnote{\url{https://spacy.io}} to extract all of the named entities and their NER labels in the passage. Then, we apply Berkeley Neural Parser \cite{kitaev-klein-2018-constituency} to parse each sentence and extract a longer constituent containing a named entity with the constraint of $\omega$\% sentence length as the final answer span. Here, we set $\omega=80$. In our experiment, we consider named entity (NE), noun phrase (NP), adjective phrase (ADJP), verb phrase (VP), and sub-clause (S) as the candidate answer types. The dataset consists of 908,511 QA pairs. We randomly sample 300,000 to initially train a QA model, 600,000 to split them into $N=$6 parts (followed the empirical results in \citet{li2020harvesting}) for the filtering phase.

\paragraph{Question Answering Model Settings.} We use BERT as the backbone of our QA model. We use Adam \cite{kingma2014adam} as the optimizer. The learning rate is 3e-5 and the batch size is 24. The max sequence length is 384 and the doc stride is 128. The discriminator is set as a one-layer network. We set $L=5,\alpha=1,\beta=1$. We use BERT-large-uncased-whole-word-masking, train the model for 2 epochs, save the checkpoint every 1,000 training steps and use the dev set to evaluate them for early stopping. Then, we continuously train the model with filtered data via the denoising filter, where $K=1$ and the substring threshold $\gamma=0.1$.

\subsection{Results}

We evaluate our model on SQuAD v1.1, NewsQA, TriviaQA, NaturalQuestions (NQ),  BioASQ and DuoRC. We compare DiverseQA with supervised approaches \cite{wang2016machine,seo2016bidirectional,devlin2018bert}, unsupervised approaches \cite{lewis2019unsupervised,li2020harvesting} and the approach using a supervised summarization dataset \cite{lyu2021improving}. We use Exact Match (EM) and F1 score as our metrics. We use the pre-processed data provided in MRQA \cite{fisch2019mrqa}.

The experimental results on six different benchmarks are shown in Table \ref{tab:compare_baselines}. The model trained on the synthetic QA data created by our DiverseQA reaches the state-of-the-art on five benchmarks, which shows the competitive performance of the proposed method in a wide range of domains. However, we find that our model underperforms on NaturalQuestions (NQ) dataset on exact match (EM). It is because the answers in this dataset are all entities, which means the model that only learns information from named entity\footnote{Although `named entity' and `entity' are different, they share the common features in most aspects.} can have good performance. After all, the model only needs to choose a proper entity from an entity set rather than extract a possible span from the whole context, which finally leads to a good EM value. But as we know, named entities (or entities) are not the only answer type in the real world. Therefore, it's unfair for our method. Because \citet{lyu2021improving} (the line under `\textit{Trained with Supervised Summarization Dataset}' in Table \ref{tab:compare_baselines}) makes use of supervised summarization dataset XSUM \cite{narayan2018don} to train the model, which is not a purely unsupervised method, we don't compare our method with it.

\subsection{Analysis}

We conduct experiments in this section to further understand our method. BERT-base-uncased is used to complete each experiment.

\subsubsection{Effects of Different Components of DiverseQA}\label{sec:effects_of_different_components_of_diverseqa}

We conduct experiments on different components of DiverseQA. A brief illustration is as follows:

\paragraph{NeAnsQA} Only extract NE as the answer type.

\paragraph{DiverseAnsQA} Take the data-extending strategy proposed to build a dataset with diverse answers.

\paragraph{RandomAnsQA} Each answer span in \textit{RandomAnsQA} dataset has the same length with that of DiverseAnsQA while the answer span is randomly extended from the original NE-based answer.

\paragraph{Adjusting Vector} It imposes the adjusting vector produced by the `Adjustor' module on the embeddings of input tokens as an augmentation instance.

\paragraph{Answer-type Discriminator} The module is to classify the adjusting vector into an answer type.

\paragraph{Top-K Filter} Apply the Top-K Filter described in Section \ref{sec:denoising-filter}.

\paragraph{Substring Filter} Apply the Substring Filter described in Section \ref{sec:denoising-filter}.

\begin{table}[t]   

    \centering
    \begin{tabular}{lrr}    
        \toprule    
        & EM & F1\\
        \midrule
        NeAnsQA & 49.2& 59.3\\
        RandomAnsQA & 48.7& 62.0\\
        \midrule
        DiverseAnsQA& 52.1& 64.0\\
        \ \ + Answer-type Classifier*& 51.3& 63.6\\
        \ \ + Adjusting Vector& 52.3& 64.0\\
        \ \ + Answer-type Discriminator& 52.9 & 64.6\\
         \ \ + Top-K  Filter & 54.7& 65.6\\
          \ \ + Substring  Filter & \textbf{55.0}& \textbf{66.2}\\
        \bottomrule
    \end{tabular}  
    \caption{\label{tab:uda-ablation}Ablations on each component of DiverseQA method on the SQuADv1.1 development set. Components below the component with `*' are not added upon the `*' component.}   
\end{table}
As shown in Table \ref{tab:uda-ablation}, the result illustrates that each component can improve the performance of our model. The difference between \textit{NeAnsQA} and \textit{DiverseAnsQA} shows that the use of our data construction strategy can greatly improve the performance of the model. It also demonstrates the importance of diverse answer types in unsupervised QA. Because answer length distribution is changed from NeAnsQA to DiverseAnsQA, we design \textit{RandomAnsQA} to get rid of this extra factor. It shows that DiverseAnsQA still outperform RandomAnsQA by a large gap, which demonstrates the effectiveness of the proposed QA data construction strategy. The result of \textit{Adjusting Vector} and \textit{Answer-type Discriminator} shows that although adding the adjusting vector can slightly improve the performance, answer-type discriminator can continuously improve the performance, showing that adding the answer-type constraint to the distribution of the adjusting vector can benefit the performance of the model. To verify the necessity of the Discriminator, we also use a simple `Answer-type Classifier' to classify input sequences into different answer types. The result shows that a simple classifier is not adequate to improve the method. In addition, the gains by adding the \textit{Top-K Filter} reveal the importance of the filtering phase in the training process. What's more, \textit{Substring Filter} can further gain the performance of the model. What's more, the results in the last row of Table \ref{tab:uda-ablation} show that the Substring Filter is useful.
\begin{table}[t]
    \centering
    \begin{tabular}{lccccc}    
        \toprule
        $\omega$\% &20\%& 40\% & 60\% &80\%& 100\%\\
        \midrule
        F1 & 63.0 & 63.6 & 65.4 & \textbf{66.2}&62.2 \\
        \bottomrule
    \end{tabular}
    \caption{\label{tab:span_extending_threshold}F1 scores of different Span Extending Thresold evaluated on SQuADv1.1 dev set.}
\end{table}
\begin{figure}[t]
    \begin{minipage}{\linewidth}
    \centering
        \includegraphics[width=.9\linewidth]{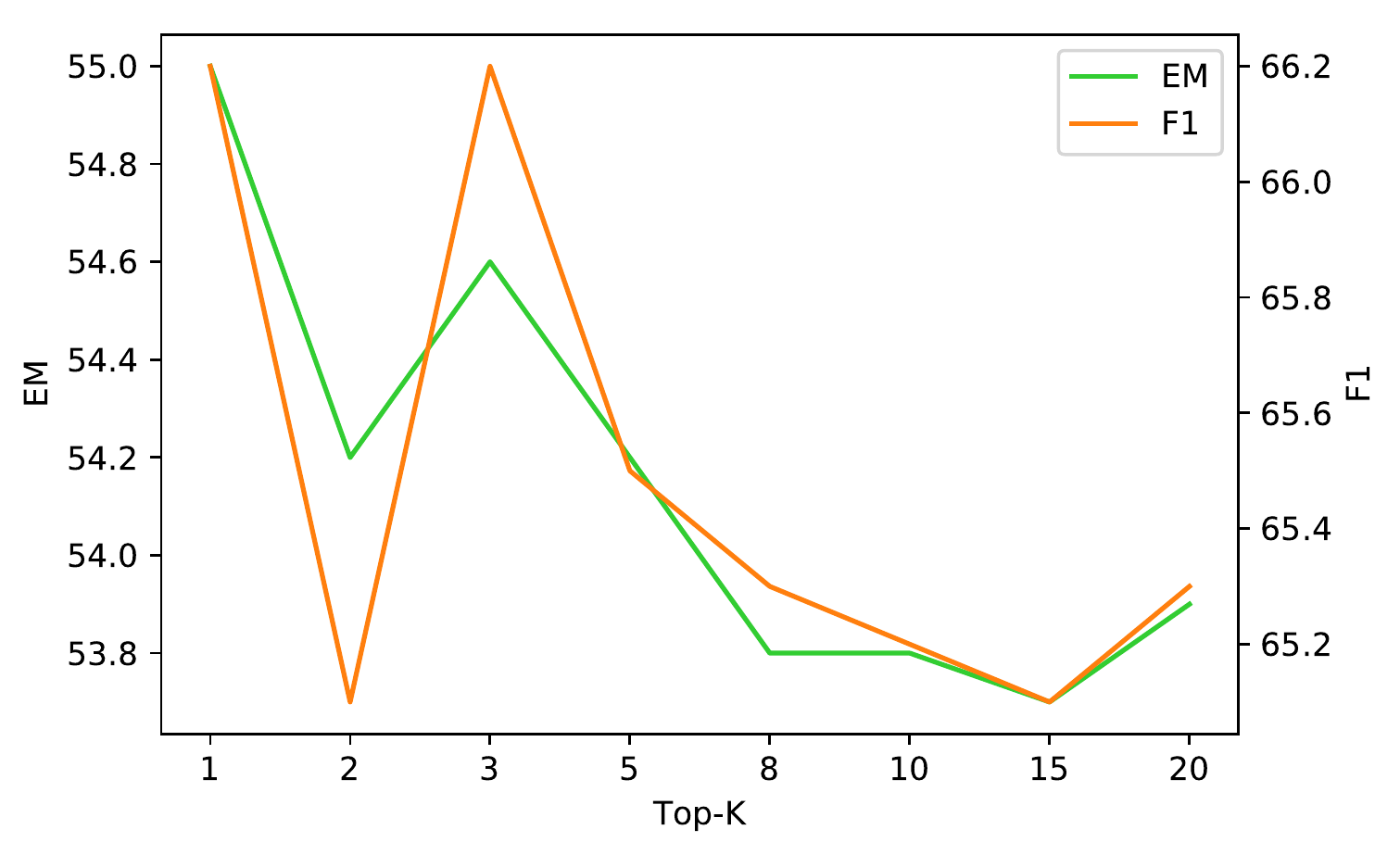}
    \end{minipage}
    \centering
    \caption{Results of using K answers with the highest predicting probabilities in the denoising filtering phase.}
    \label{fig:filter}
\end{figure}
\begin{table}[t]
    \centering
    \begin{tabular}{lcccccc}    
        \toprule
        $\gamma$ & 0.0 & 0.1 & 0.2& 0.4 & 0.6 &0.8\\
        \midrule
        F1 & 65.2 & \textbf{66.2} & 65.2 & 65.2 & 64.9 & 65.0 \\
        \bottomrule
    \end{tabular}
    \caption{\label{tab:substring_filter}F1 scores of different Substring Filter threshold evaluated on SQuADv1.1 dev set.}
\end{table}

\subsubsection{Effects of Span Extending Threshold}
We experiment with several Span Extending Threshold to construct the synthetic dataset. Table \ref{tab:span_extending_threshold} shows that the optimal value is 80\%. It illustrates that neither a too strict nor a too loose span extending condition can produce a high quality dataset.

\vspace{-0.3cm}
\subsubsection{Effects of Denoising Filter}
\begin{table}[t]
    \centering
    \begin{tabular}{lccc}    
        \toprule
         & No Filter& Refine & Denoise\\
        \midrule
        RefQA & 49.2/59.3 & 52.0/62.6 & 53.3/62.2\\
        DiverseQA & \textbf{52.1/64.0} & \textbf{52.1/64.0}* & \textbf{55.0/66.2}\\
        \bottomrule
    \end{tabular}
    \caption{\label{tab:filter_comparison}EM/F1 scores of RefQA and DiverseQA with different continuously training strategy. ``Refine'' denotes the Refining phase in RefQA. ``Denoise'' denotes the proposed Denoising Filter. `*' means that the model cannot make any improvement on the specific continuous training strategy.}
\end{table}

We conduct experiments on the K answers with the highest predicting probabilities as well as Substring Filter threshold $\gamma$, which will be used in the continuing training.
Figure \ref{fig:filter} shows that when K=1, the performance reaches the best. It means that when the model-predicted answer with high probability matches the span extracted via our extending strategy, the corresponding QA pair is probably of high quality and can be further reused to improve the performance of the model. Table \ref{tab:substring_filter} shows that when $\gamma$=0.1, the produced NE-based QA pairs with predictions of substring can be beneficial.

Furthermore, we make comparison between the proposed Denoising Filter and the Refining Phase \cite{li2020harvesting}. Table \ref{tab:filter_comparison} shows that firstly, DiverseQA largely outperforms RefQA without any filter. Secondly, the proposed Denoising Filter can be effectively used in the second phase of both RefQA and DiverseQA, showing its strong adaptation to different methods, while Refining Phase can only make improvement in RefQA. Thirdly, although the F1 score of RefQA with Denoising Filter is slightly lower than that of Refining Phase, it makes large improvement on the EM metric, demonstating the advantage of the proposed Denoising Filter.

\subsubsection{Effects of Diverse Answer Types}
\label{sec:effects_of_diverse_answer_types}
\begin{table*}[t]   
\fontsize{10.2}{11}\selectfont
    \centering
    \begin{tabular}{lccccccc}    
        \toprule
        & {NE} & {NP} & {ADJP} & {VP} & {S} & {Others} &
        {Overall}\\
        Models & EM/F1  & EM/F1  & EM/F1  & EM/F1  & EM/F1 & EM/F1 & EM/F1\\
        \midrule
        RefQA & 67.0/75.4& 47.9/60.4& \textbf{31.9}/41.4& 4.1/18.3& 22.7/35.0 & 36.7/52.2 & 52.0/62.6\\
        \midrule
        RandomAnsQA & 58.8/70.7& 43.2/58.8& 25.0/39.7& 9.4/25.6& 21.4/39.0 & 36.8/53.4 & 48.7/62.0\\
        DiverseAnsQA & 62.3/72.4& 47.2/61.3& 26.5/39.3& 13.2/29.3& 22.9/41.0 & 38.5/54.6& 52.1/64.0\\
        \midrule
        DiverseQA & 67.5/75.7& \textbf{49.1}/\textbf{62.8}& 27.9/\textbf{42.5}& \textbf{13.3}/\textbf{30.4}& \textbf{29.1/42.6} & \textbf{41.4}/\textbf{57.2} & \textbf{55.0}/\textbf{66.2}\\
        w/o Extension & \textbf{71.5/77.3}& 48.5/60.2& 24.5/35.1& 1.6/12.1& 21.9/30.6 & 34.4/47.0 & 52.0/60.6\\
        w/o Discriminator & 66.9/74.7& 47.6/60.1& 28.4/40.5& 9.7/28.0& 26.3/40.9 & 40.4/54.7 & 53.6/63.9\\
        \bottomrule
    \end{tabular}  
    \caption{\label{tab:type_generalize}EM/F1 scores of different models on the SQuADv1.1 dev set. Columns are each dev set associated with the type of answer. `Others' means other types of answers out of our consideration. `Extension' represents our answer span extending module. `Discriminator' denotes the proposed Answer-type Discriminator.}
\end{table*}
\begin{table}[t]   
\fontsize{10}{10.5}\selectfont
    \centering
    \begin{tabular}{p{.44\linewidth}p{.44\linewidth}}
        \toprule
        \multicolumn{2}{p{.9\linewidth}}{\textbf{Context:}  The Town of Estill is located in the southern half of Hampton County .}\\
        \midrule
        \textbf{Question:} Where of the southern half in The town of Estill is located &\textbf{ Question:} Where of the town Estill\\
        \textbf{The answer in RefQA:} Hampton County&\textbf{Our answer:} is located in the southern
half of Hampton County\\
        \bottomrule
    \end{tabular}
    \caption{\label{tab:examples}An example of RefQA (left half) and our QA pairs (right half).}
\end{table}

We experiment on the SQuADv1.1 dev set partitioned by answer types. For a fair comparison, the re-implement of RefQA shares the raw text (from which we generate QA pairs and train the model) and the random seed with ours. 

Although \citet{li2020harvesting} claims that their refine phase can generate diverse answers, the result in Table \ref{tab:type_generalize} shows that DiverseQA outperforms the performance of RefQA on nearly every kind of answer type, especially on VP and S by large margins. It is because the refine phase proposed in RefQA heavily relies on the QA model training with purely NE-based QA pairs. Therefore, the trained model could probably generate certain variants of NEs, which are short and used to continuously train the model, leaving relatively long answer spans like VPs and Ss out. Besides, we observe that the performance of DiverseQA is lower than RefQA on ADJP in the exact match (EM). It is because the ADJP merely accounts for 0.3\% (in Appendix \ref{appendix:answer_type_distribution}) in the whole synthetic dataset generated by DiverseQA, while the refining phase in RefQA may probably generate many variants of NEs, which could be ADJPs\footnote{For instance, in `\textit{he was still being paid more than \$ 10,000 as a legal advisor to the Chicago}', span `\textit{10,000}' is a named entity with NER label `MONEY' while `\textit{\$ 10,000}' is a variant of the named entity as well as an ADJP.}. Consequently, the model in RefQA might train with too many ADJP-like QA instances and prefer to choose the ADJPs as answer types, leading to the result in Table \ref{tab:type_generalize}.
Since in our method, we only consider NE, NP, ADJP, VP, S as our answer spans, it's important to know how well our method performs on other types of answers out of consideration. The result in the `\textit{Others}' column of Table \ref{tab:type_generalize} demonstrates that the proposed model 
can generalize to other unseen answer types and outperform RefQA by a large gap.

In addition, we ablate the proposed span extending strategy and Answer-type Discriminator separately to explore the performance of the two components. The result in the last two rows of Table \ref{tab:type_generalize} shows that the two components can both contribute to the performance of model on most of answer types. Besides, we notice that the performance of ``w/o Extension'' on the NE obtains the best. It is because that under this setting, the model will only learn from NE-based QA pairs. Therefore, it prefers to choose a NE from the context, which can gain its performance on NE-based QA pairs.

What's more, we also show an example to make the comparison between the QA pair in DiverseQA and that of RefQA in Table \ref{tab:examples}. In the example, the question `Where of the town Estill' needs more reasoning process to be answered correctly than the question generated in RefQA. As in the proposed method, when the generated answer becomes longer, the generated question becomes shorter. Therefore, it's also necessary to keep the question and answer length unchanged while exploring the effectiveness of answer diversity. In Table \ref{tab:type_generalize}, the results of RandomAnsQA and DiverseAnsQA demonstrate the usefulness of the proposed answer extension method when both the question and answer length distributions are unchanged. More examples are shown in Appendix \ref{appendix:case_study}.

\subsubsection{Effects of Answer Length Distribution}
In the proposed method, the first module is to extend the span into a longer constituent. Although it indeed introduces new answer types and gains performance, answer length distribution also changes simultaneously. Therefore, we further explore how the drift of answer length distribution affects the model's performance. 

\begin{figure}[t]
    \begin{minipage}{\linewidth}
    \centering
        \includegraphics[width=.9\linewidth]{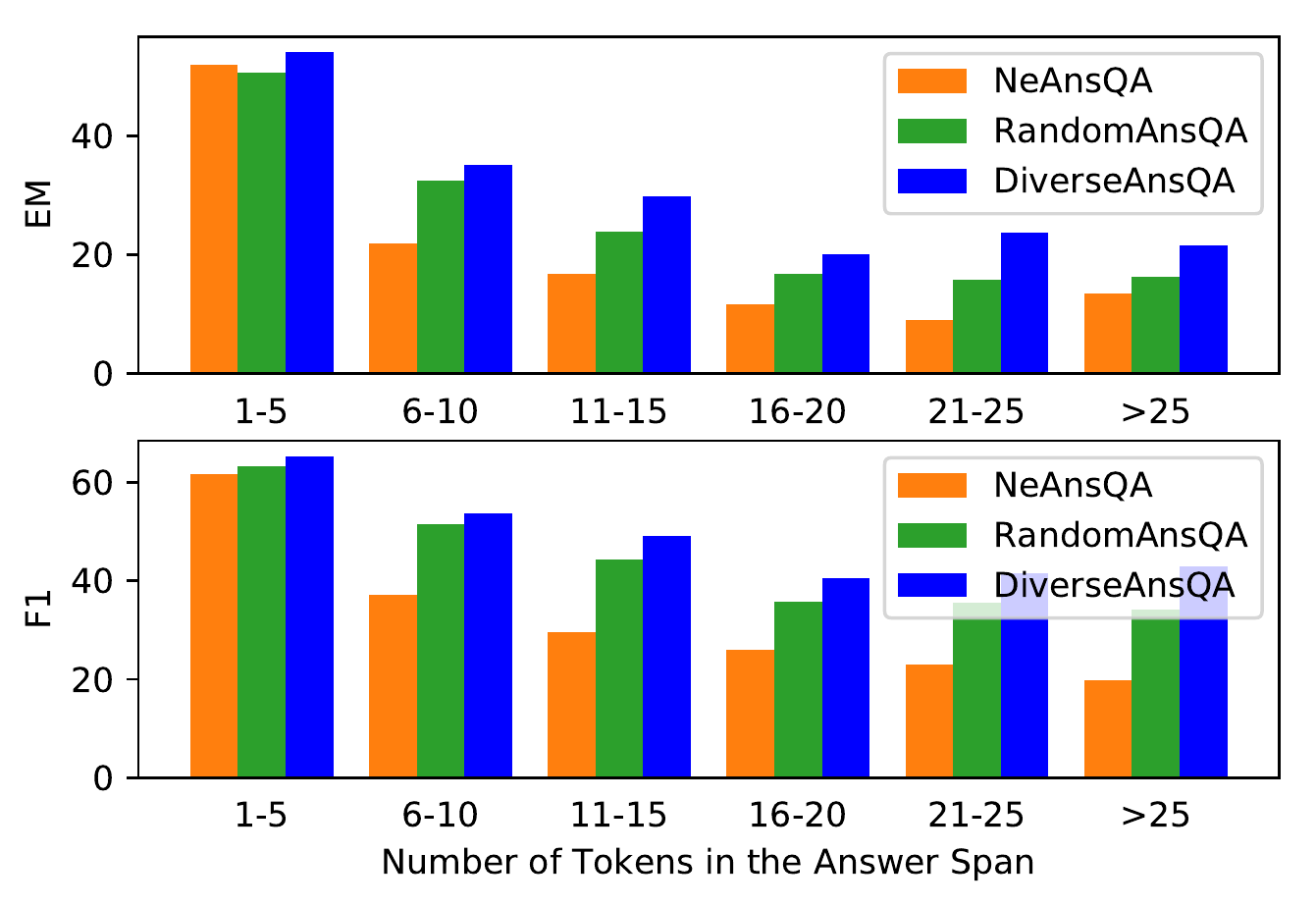}\\
        \vspace{0.02cm}
    \end{minipage}
    \centering
    \vspace{-0.2cm}
    \caption{\label{fig:answer_length_performance}Comparison among \textit{NeAnsQA}, \textit{RandomAnsQA} and \textit{DiverseAnsQA} (Section \ref{sec:effects_of_different_components_of_diverseqa}). SQuADv1.1 dev set is partitioned by answer length for evaluation.}
\end{figure}

As shown in Figure \ref{fig:answer_length_performance}, the performance of NeAnsQA is lower than RandomAnsQA and DiverseAnsQA. When the answer length becomes larger (>10 tokens), the gap is even larger. This demonstrates that model trained only with NE-based QA pairs lacks the ability to handle QA pairs with long answers. Besides, it can be found that the performance of RandomAnsQA is lower than DiverseAnsQA on each answer length. It means that without the influence of answer distribution, the proposed span extending strategy can generate higher quality instances than that of randomly extending strategy.

\begin{figure}[t]
    \begin{minipage}{\linewidth}
    \centering
        \includegraphics[width=.9\linewidth]{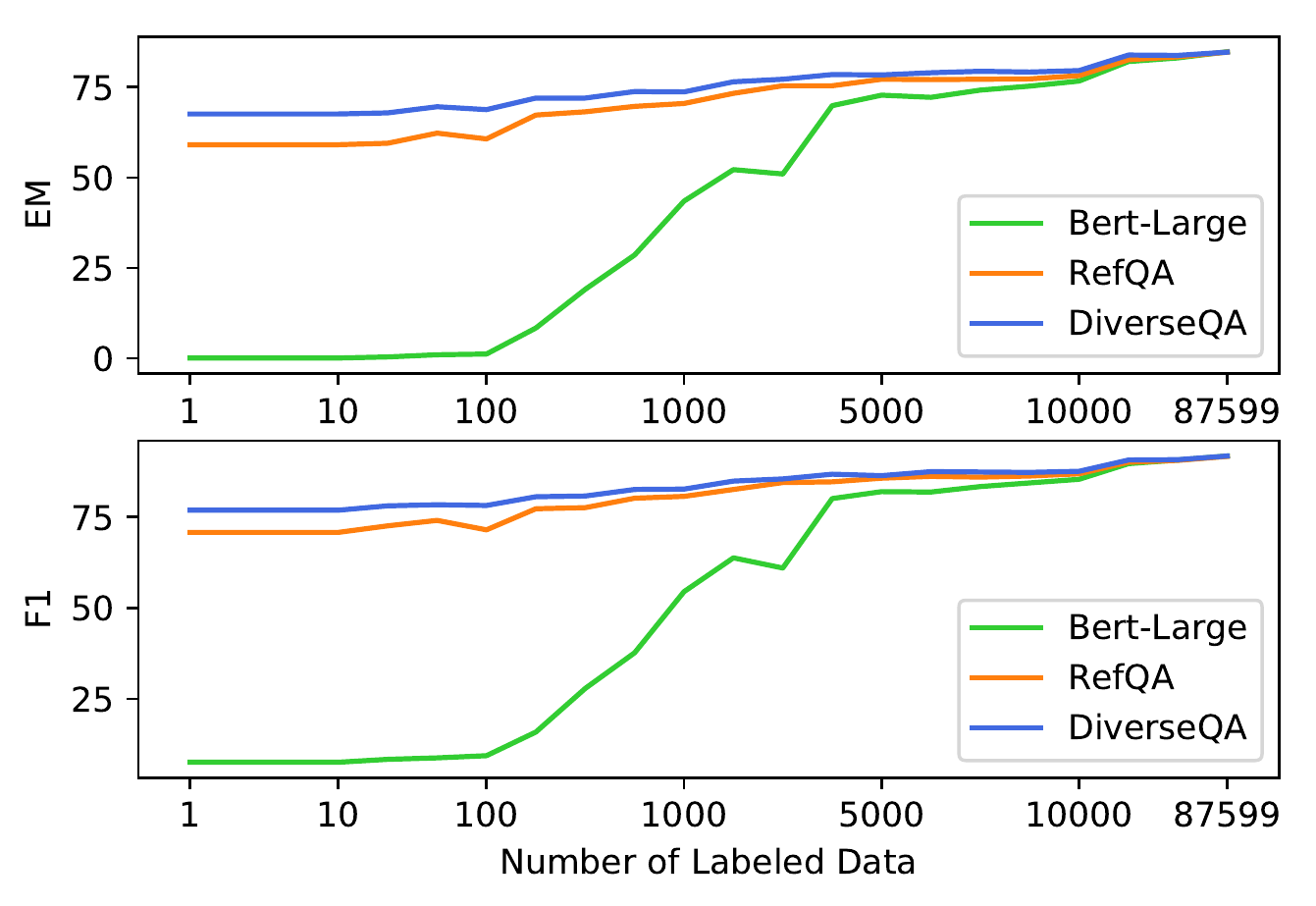}\\
        \vspace{0.02cm}
    \end{minipage}
    \centering
    \vspace{-0.2cm}
    \caption{Few-shot results (EM and F1) using different sizes of SQuADv1.1 training data, comparing among DiverseQA, RefQA \protect\cite{li2020harvesting} and BERT-large-uncased-whole-word-masking.}
    \label{fig:few-shot}
\end{figure}
\vspace{-0.1cm}

\subsection{Few-Shot Learning}
We conduct experiments in the few-shot learning setting. For a fair comparison, we use our best-trained model without the data augmentation component, our re-implementation of RefQA, and a BERT model, all of which use the BERT-large-whole-word-masking as the backbone.

Figure \ref{fig:few-shot} shows that our model reaches the best result trained on various sizes of supervised data ranging from 1 to 50,000, demonstrating the strong ability of our method in the low-resource scenario. 

\section{Conclusion}
We propose DiverseQA, an unsupervised QA method comprising a synthetic QA dataset with diverse answers, an answer-type-dependent data augmentation process via adversarial training, and a denoising filter to improve the performance of a QA model. Our method reaches the state-of-the-art on five benchmarks and shows strong performance in the few-shot learning setting.



\bibliography{anthology,custom}
\bibliographystyle{acl_natbib}

\newpage
\appendix

\section*{Appendix}
\section{The Distribution of Extracted Answers}
\label{appendix:answer_type_distribution}
\begin{table}[h!]
  \centering
  \begin{tabular}{lrr}
    \toprule
    & \#Instances & \%Frequency\\
    \midrule
    NE &  716,716 & 78.9\%\\
    NP &  161,178 & 17.7\%\\
    ADJP &  2,563 & 0.3\%\\
    VP &  23,420 &  2.6\%\\
    S & 4,634 &  0.5\%\\
    \bottomrule
  \end{tabular}
  \caption{The statistics of extracted answer types in the synthetic dataset constructed using DiverseQA.}
  \label{tab:extracted-answers}
\end{table}
\label{sec:our-answer-distribution}

The distribution of our extracted answer types (NE, NP, ADJP, VP, S) is shown in Table \ref{tab:extracted-answers}. The frequencies of each answer type are used in the modified predicting distribution described in Section \ref{sec:embedding-adjusting}. Since previous works consider named entities as the only answer type (i.e. NE acounts for 100\% of all the answers), we don't show the answer type distribution of them. Besides, as the answer span generated by the trained QA model in the refining phase of \citet{li2020harvesting} fails to follow the rule of constituent parsing, we cannot obtain its answer type distribution in terms of sentence constituents.

\section{Answer Length Distribution}
\begin{figure}[h]
    \begin{minipage}{\linewidth}
    \centering
        \includegraphics[width=.9\linewidth]{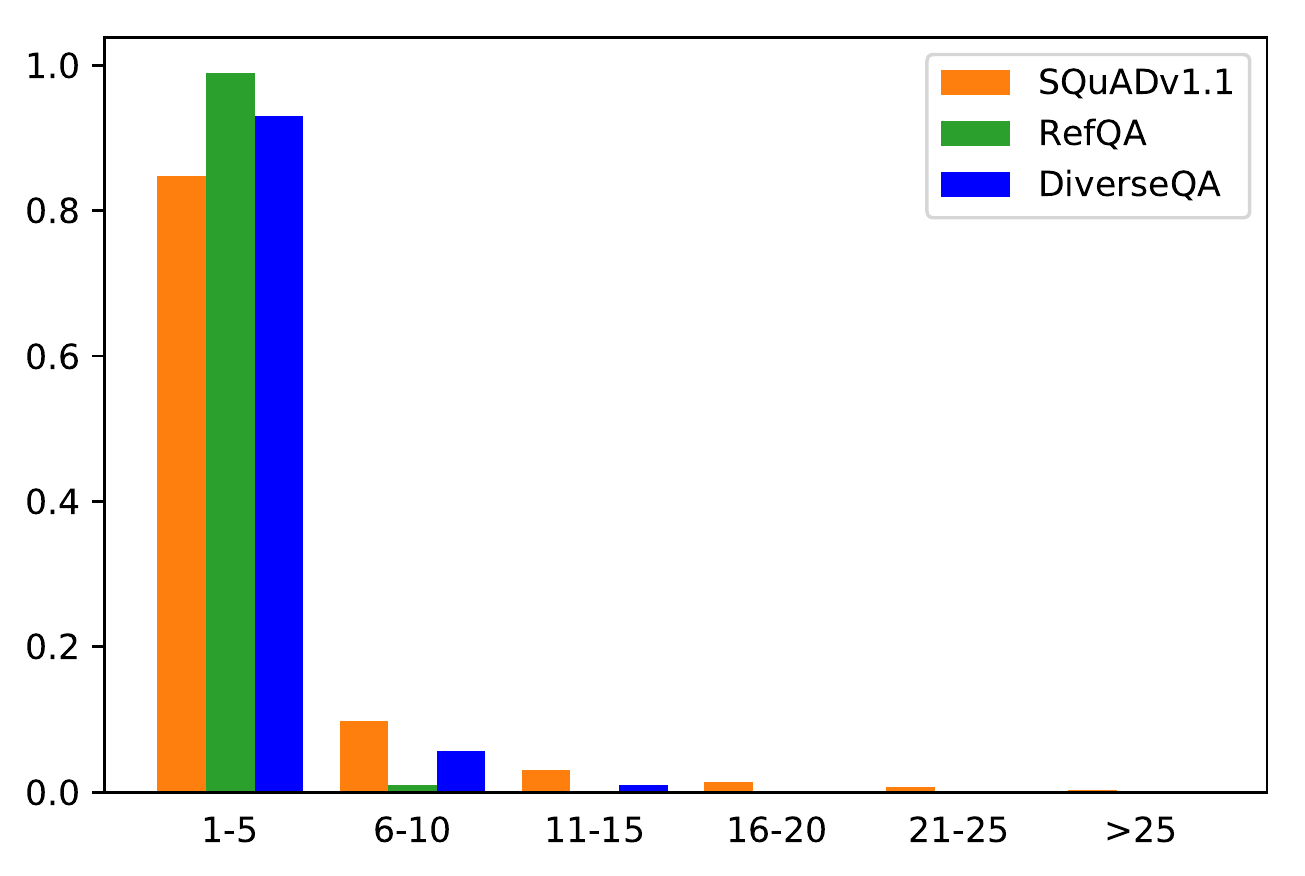}\\
        \vspace{0.02cm}
    \end{minipage}
    \centering
    \vspace{-0.2cm}
    \caption{\label{fig:answer_length_distribution}Answer length distributions.}
\end{figure}

We randomly sample 10,000 instances from SQuADv1.1, the dataset of RefQA, and the dataset of DiverseQA respectively, and count the answer length distributions, which are shown in Figure \ref{fig:answer_length_distribution}. It shows that firstly, the proposed DiverseQA method can construct dataset with a similar answer length distribution to the annotated dataset SQuADv1.1. Secondly, the dataset only with NE-based QA pairs (like RefQA) is not able to cover long form answers (it accounts for 0\% QA pairs in the answer length ranges `11-15', `21-25' and `>25')\footnote{The dataset of DiverseQA accounts for 0.16\%, 0.13\% and 0.03\% in the ranges `16-20', `21-25' and `>25'. And it is not displayed properly in Figure \ref{fig:answer_length_distribution}.}, which is harmful to the performance of model on long answers.

\section{Algorithms}
We describe the whole training procedure as follows:

\begin{algorithm}[h]
  \SetAlgoLined
  \KwData{Synthetic QA dataset $\mathcal{D}$, a BERT model $\mathcal{M}$ with answer-type dependent data augmentation module, a denoising filter composed of a Top-K filter and a Substring Filter with threshold $\gamma$.}
  \KwResult{A fine-tuned model $\mathcal{M}'$.}
  Split $\mathcal{D}$ into $\mathcal{D}_I$ and $\mathcal{D}_F$\;
  Fine-tune $\mathcal{M}$ with $\mathcal{D}_I$\;
  Split $\mathcal{D}_F$ equally into $\{\mathcal{D}_{F_i}\}_{i=1}^N$\;
  \For{$i\leftarrow 1$ \KwTo $N$}{
    $\mathcal{S}\leftarrow \emptyset$\;
    \ForEach{\textup{element} $e$ \textup{in} $\mathcal{D}_{F_i}$}{
        Obtain the probability $p_e$ using $\mathcal{M}$\;
        \If{$p_e$ \textup{is in the Top-K highest probabilities}}{$\mathcal{S} \leftarrow \mathcal{S} \cup \{e\}$}
        \If{\textup{the answer of} $e$ \textup{is a subtring of the extracted one} $\mathbf{and}$ $p_e \ge \gamma$}{$\mathcal{S} \leftarrow \mathcal{S} \cup \{e\}$}
    }
    Fine-tune model $M$ with dataset $\mathcal{S}$\;
  }
  \caption{Training Procedure}
\end{algorithm}

\section{Generated QA Instances}
\label{appendix:case_study}
\begin{table*}[t]   
    \centering
    \begin{tabular}{p{.45\textwidth}p{.45\textwidth}}
        \toprule
        \multicolumn{2}{p{.9\textwidth}}{\textbf{Context:} Joss Whedon has endorsed Mitt Romney ( in a way ) and now `` The Simpsons '' Mr. Burns , owner of Springfield 's nuclear power plant , titan of corporate capitalism and honcho in the Springfield Republican Party , has come out with his own backhanded endorsement of the Republican nominee .}\\
        \hdashline
        \textbf{Question:}Who As the chief of ``Springfield Republican Party'' endorsed Mitt Romney in 2012 US Presidential Election . 
        &\textbf{Question:}  Who As the chief of endorsed Mitt Romney in 2012 US Presidential Election .  \\
        \textbf{Answer:} Burns
        &\textbf{Answer:} Mr. Burns\\
        \midrule
        \multicolumn{2}{p{.9\textwidth}}{\textbf{Context:} In 1931 , the Singers gave the Museum of Fine Arts to the community along with a substantial collection of American and European art .}\\
        \hdashline
        \textbf{Question:} Who American and art of a substantial collection with along the Singers gave the Museum of Fine Arts to the community . &\textbf{Question:}  Who art of a substantial collection with along the Singers gave the Museum of Fine Arts to the community .\\
        \textbf{Answer:} European&\textbf{Answer:} American and European\\
        \midrule
        \multicolumn{2}{p{.9\textwidth}}{\textbf{Context:} Why do the new prequels sometimes contradict the history set forth in THE DUNE ENCYCLOPEDIA compiled by,
            Dr. Willis E. McNelly ?,
             THE DUNE ENCYCLOPEDIA reflects an alternate `` DUNE universe '' which did not necessarily represent the `` canon '' created by Frank Herbert ."}\\
        \hdashline
        \textbf{Question:}Who by written to accompany the ``Dune'' books &\textbf{Question:}  Who by written to accompany the ``Dune'' books \\
        \textbf{Answer:} Willis E. McNelly&\textbf{Answer:} Dr. Willis E. McNelly\\
        \midrule
        \multicolumn{2}{p{.9\textwidth}}{\textbf{Context:}GDP per capita in the city increased by 2.4 per cent and employment by 4.7 per cent compared to the previous year .}\\
        \hdashline
        \textbf{Question:} How much by GDP per capita in the city increased &\textbf{Question:}  How much in GDP per capita\\
        \textbf{Answer:} 2.4 per cent&\textbf{Answer:} increased by 2.4 per cent\\
        \midrule
        \multicolumn{2}{p{.9\textwidth}}{\textbf{Context:}Under the agreement , Washington Mutual will buy HomeSide 's parent company , SR Investment Inc. , and its assets which include mortgage servicing rights on a mortgage portfolio worth about \$ 131 billion .}\\
        \hdashline
        \textbf{Question:} How many worth for \$ 1.3 billion in cash and the assumption of \$ 735 million in debt a mortgage portfolio on the mortgage servicing rights included &\textbf{Question:}  How much on the mortgage servicing rights included for \$ 1.3 billion in cash and the assumption of \$ 735 million in debt\\
        \textbf{Answer:} about \$ 131 billion&\textbf{Answer:} worth about \$ 131 billion\\
        \bottomrule
    \end{tabular}

    \caption{\label{tab:examples-appendix}Examples of the QA pairs of RefQA (left half) and that of DiverseQA (right half).}
\end{table*}

\end{document}